# NeuralNetwork Based 3D Surface Reconstruction

Vincy Joseph

Computer Department
Thadomal Shahani Engineering College,
Bandra, Mumbai, India.
vincyej@gmail.com

Shalini Bhatia

Computer Department
Thadomal Shahani Engineering College,
Bandra, Mumbai, India.
shalini.tsec@gmail.com

*Abstract*—This paper proposes a novel neural-network-based adaptive hybrid-reflectance three-dimensional (3-D) surface reconstruction model. The neural network combines the diffuse and specular components into a hybrid model. The proposed model considers the characteristics of each point and the variant albedo to prevent the reconstructed surface from being distorted. The neural network inputs are the pixel values of the two-dimensional images to be reconstructed. The normal vectors of the surface can then be obtained from the output of the neural network after supervised learning, where the illuminant direction does not have to be known in advance. Finally, the obtained normal vectors can be applied to integration method when reconstructing 3-D objects. Facial images were used for training in the proposed approach

## I. INTRODUCTION

Shape recovery is a classical computer vision problem. The objective of shape recovery is to obtain a three-dimensional (3-D) scene description from one or more two-dimensional (2-D) images. Shape recovery from shading (SFS) is a computer vision approach, which reconstructs 3-D shape of an object from its shading variation in 2-D images. When a point light source illuminates an object, they appear with different brightness, since the normal vectors corresponding to different parts of the object's surface are different. The spatial variation of brightness, referred to as *shading,* is used to estimate the orientation of surface and then calculate the depth map of the object

## II. DIFFERENT APPROACHES FOR RECONSTRUCTION

### A. Lambertian Model

A successful reflectance model for surface reconstructions of objects should combine both diffuse and specular components [1]. The Lambertian model describes the relationship between surface normal and light source direction by assuming that the surface reflection is due diffuse reflection only. This model ignores specular component.

### B. Hybrid Reflectance Model

A novel hybrid approach generalizes the reflectance model by considering both diffuse component and specular component. This model does not require the viewing direction and the light source direction and yields better shape recovery than previous approaches.

A hybrid approach uses two self-learning neural networks to generalize the reflectance model by modeling the pure Lambertian surface and the specular component of the non-Lambertian surface, respectively. However, the hybrid approach still has two drawbacks:
1) The albedo of the surface is disregarded or regarded as constant, distorting the recovered shape.
2) The combination ratio between diffuse and specular components is regarded as constant, which is determined by trial and error.

### C. Neural Network Based Hybrid Reflectance Model

This model intelligently integrates both reflection components. The pure diffuse and specular reflection components are both formed by similar feed-forward neural network structures. A supervised learning algorithm is applied to produce the normal vectors of the surface for reconstruction. The proposed approach estimates the illuminant direction, viewing direction, and normal vectors of object surfaces for reconstruction after training. The 3-D surface can also be reconstructed using integration methods.

## III. DESCRIPTION

Fig. 1 shows the schematic block diagram of the proposed adaptive hybrid-reflectance model, which consists of the diffuse and specular components. This diagram is used to describe the characteristics of diffuse and specular components of adaptive hybrid-reflectance model by two neural networks with similar structures. The composite intensity $R_{hybrid}$ is obtained by combining diffuse intensity $R_d$ and the specular intensity $R_s$ based on the adaptive weights $\lambda_d(x,y)$ and $\lambda_s(x,y)$. The system inputs are the 2-D image intensities of each point, and the outputs are the learned reflectance map.

Fig. 2 shows the framework of the proposed symmetric neural network which simulates the diffuse reflection model. The input/output pairs of the network are arranged like a mirror in the center layer, where the number of input nodes equals the number of output nodes, making it a symmetric neural network.

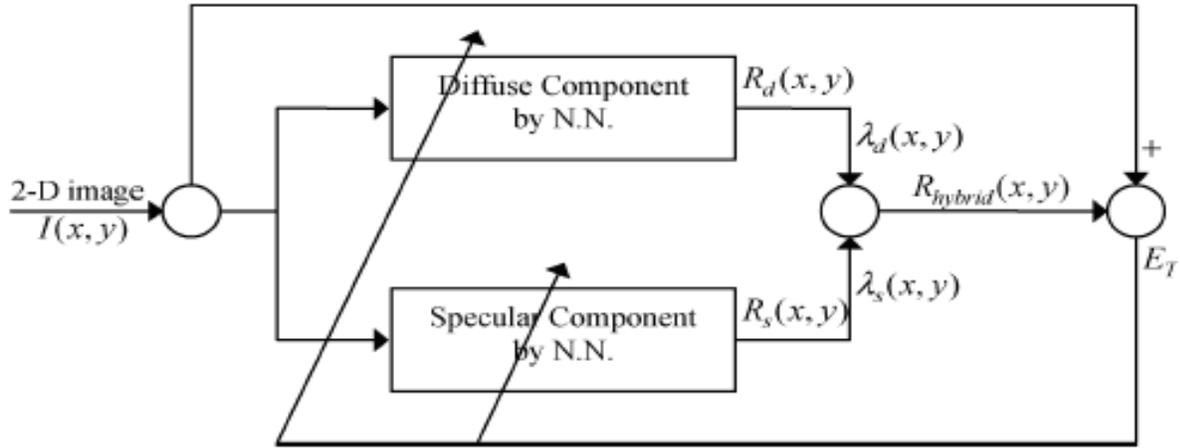

Figure 1 Block diagram of the proposed adaptive hybrid-reflectance model

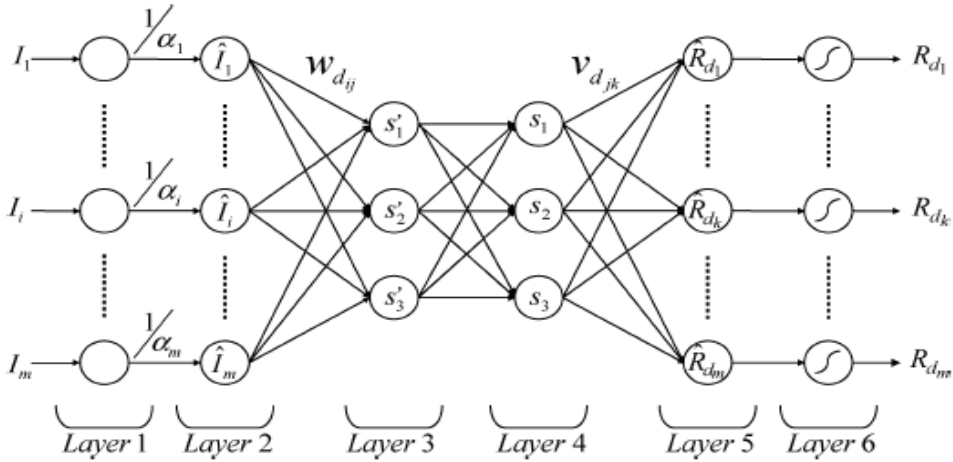

Figure 2 Framework of the proposed symmetric neural network for diffuse reflection model

The light source direction and the normal vector from the input 2-D images in the left side of the symmetric neural network are separated and then combined inversely to generate the reflectance map for diffuse reflection in the right side of the network. The function of each layer is discussed in detail below.

### A. Function of Layers

*Layer 1:* This layer normalizes the intensity values of the input images. Node $I_i$ denotes the $i^{th}$ pixel of the 2-D image and $m$ denotes the number of total pixels of the image. That is

$$f_i = I_i, \quad i = 1,...., \ m$$
$$a_i^{(1)} = f_i, \quad i = 1,...., m \tag{1}$$

*Layer 2:* This layer adjusts the intensity of the input 2-D image with corresponding albedo value.

$$f_i = \frac{I_i}{\alpha_i}, \quad i = 1,...., m$$

*Layer 3:* The purpose of Layer 3 is to separate the light source direction from the 2-D image. The light source directions of this layer are not normalized.

$$f_j = \sum_{i=1}^{m} \hat{I}_i \omega_{d_{i,j}}, \ i = 1,...,m \ \ j = 1,2,3 \tag{3}$$

$$s'_j = a_j^{(3)} = f_j, \ j = 1,2,3$$

*Layer 4:* The nodes of this layer represent the unit light source. Equation (4) is used to normalize the non-normalized light source direction obtained in Layer 3.

$$f_j = \frac{1}{\sqrt{s_1'^2 + s_2'^2 + s_3'^2}} \quad j = 1,2,3 \qquad (4)$$

$$s_j' = a_j^{(4)} = f_j . s_j' = \frac{s_j'}{\sqrt{s_1'^2 + s_2'^2 + s_3'^2}} \qquad (5)$$

*Layer 5:* Layer 5 combines the light source direction **s** and normal vectors of the surface to generate the diffuse reflection reflectance map.

$$f_k = \sum_{j=1}^{3} s_j \upsilon_{d,j,k}, \qquad k = 1,...,m$$

$$\hat{R}_{d_k} = a_j^{(5)} = f_k, \qquad k = 1,...m \qquad (6)$$

*Layer 6:* This layer transfers the non-normalized reflectance map of diffuse reflection obtained in Layer 5 into the interval [0,1].

$$f_k = \hat{R}_{d_k}, \ k = 1,...,m$$

$$\hat{R}_{d_k} = a_j^{(6)}$$

$$= \frac{255 \left( f_k - \min \left( \hat{R}_d \right) \right)}{\max(\hat{R}_d) - \min(\hat{R}_d)}$$

$$= \frac{255 \left( \hat{R}_{d_k} - \min \left( \hat{R}_d \right) \right)}{\max(\hat{R}_d) - \min(\hat{R}_d)}$$

$$k = 1,...,m \qquad (7)$$

where $\left( \hat{R}_{d_1}, \hat{R}_{d_2},..., \hat{R}_{d_m} \right)^T$ and the link weights between Layers 5 and 6 are unity.

Similar to the diffuse reflection model, a symmetric neural network is used to simulate the specular component in the hybrid-reflectance model. The major differences between these two networks are the node representation in Layers 3 and 4 and the active function of Layer 5. Through the supervised learning algorithm derived in the following section, the normal surface vectors can be obtained automatically.[3] Then, integration methods can be used to obtain the depth information for reconstructing the 3-D surface of an object by the obtained normal vectors[4].

### B. Training Algorithm

Back-propagation learning is employed for supervised training of the proposed model to minimize the error function defined as

$$E_T = \sum_{i=1}^{m} \left( R_{hybrid_i} - D_i \right)^2 \qquad (8)$$

where m denotes the number of total pixels of the 2-D image, $R_i$ denotes the i th output of the neural network, and $D_i$ denotes the i th desired output equal to the i th intensity of the original 2-D image. For each 2-D image, starting at the input nodes, a forward pass is used to calculate the activity levels of all the nodes in the network to obtain the output. Then, starting at the output nodes, a backward pass is used to calculate $\partial E_T / \partial \omega$, where ω denotes the adjustable parameters in the network. The general parameter update rule is given by

$$\omega(t+1) = \omega(t) + \Delta \omega(t)$$

$$= \omega(t) + \eta \left( -\frac{\partial E_T}{\partial \omega(t)} \right) \qquad (9)$$

The details of the learning rules corresponding to each adjustable parameter are given below.

### C. The Output Layer

The combination ratio for each point $\lambda_{dk}(t)$ and $\lambda_{sk}(t)$ is calculated iteratively by

$$\lambda_{dk}(t+1) = \lambda_{dk}(t) + \Delta \lambda_{dk}(t)$$

$$= \lambda_{dk}(t) + 2\eta(D_k(t) - R_{hybridk}(t))R_{dk}(t)$$

$$k = 1,...,m$$
$$(10)$$

$$\lambda_{sk}(t+1) = \lambda_{sk}(t) + \Delta \lambda_{sk}(t)$$

$$= \lambda_{sk}(t) + 2\eta(D_k(t) - R_{hybridk}(t))R_{sk}(t)$$

$$k = 1,...,m$$
$$(11)$$

where $D_k(t)$ denotes the *k*th desired output; $R_{hybridk}(t)$ denotes the *k*th system output; $R_{dk}(t)$ denotes the *k*th diffuse intensity obtained from the diffuse subnetwork; $R_{sk}(t)$ denotes the *k*th specular intensity obtained from the specular subnetwork;m denotes the total number of pixels in a 2-D image, and η denotes the learning rate of the neural network. For a gray image, the intensity value of a pixel is in the interval [0, 1]. To prevent the intensity value of $R_{hybrid}$ from exceeding the interval [0, 1], then the rule $\lambda_d + \lambda_s = 1$ where $\lambda_d > 0$ and $\lambda_s > 0$, must be enforced. Therefore, the combination ratio $\lambda_{dk}$ and $\lambda_{sk}$ is normalized by

$$\lambda_{dk}(t+1) = \frac{\lambda_{dk}(t+1)}{\lambda_{dk}(t+1) + \lambda_{sk}(t+1)} \quad k = 1,...,m$$

$$\lambda_{sk}(t+1) = \frac{\lambda_{sk}(t+1)}{\lambda_{dk}(t+1) + \lambda_{sk}(t+1)} \quad k = 1,...,m \quad (12)$$

### D. Subnetworks

The normal vector calculated from the subnetwork corresponding to the diffuse component is denoted as $n_{dk} = (\upsilon_{d1k}, \upsilon_{d2k}, \upsilon_{d3k})$ for the $k$th point on the surface, and the normal vector calculated from the subnetwork corresponding to the specular component is denoted as $n_{sk} = (\upsilon_{s1k}, \upsilon_{s2k}, \upsilon_{s3k})$ for the $k$th point.

The normal vectors $n_{dk}$ and $n_{sk}$ are updated iteratively using the gradient method as

$$\upsilon_{djk}(t+1) = \upsilon_{djk}(t) + \Delta\upsilon_{djk}(t)$$

$$= \upsilon_{djk}(t) + 2\eta s_j(t)(D_k(t) - R_{hybridk}(t))$$

$$j = 1,2,3 \quad (13)$$

$$\upsilon_{sjk}(t+1) = \upsilon_{sjk}(t) + \Delta\upsilon_{sjk}(t)$$

$$= \upsilon_{sjk}(t) + 2\eta r h_j(t)(D_k(t) - R_{hybridk}(t))$$

$$j = 1,2,3 \quad (14)$$

where $s_j(t)$ denotes the $j$th element of illuminant direction **s** ; $h_j(t)$ denotes the $j$th element of the halfway vector , and $r$ denotes the degree of the specular equation. The updated $\upsilon_{djk}$ and $\upsilon_{sjk}$ should be normalized as follows:

$$\upsilon_{djk}(t+1) = \frac{\upsilon_{djk}(t+1)}{\|n_{dk}(t+1)\|}$$

$$\upsilon_{sjk}(t+1) = \frac{\upsilon_{sjk}(t+1)}{\|n_{sk}(t+1)\|} \quad j = 1,2,3 \quad (15)$$

To obtain the reasonable normal vectors of the surface from the adaptive hybrid-reflectance model, $n_{dk}$ and $n_{sk}$ are composed from the hybrid normal vector $n_k$ of the surface on the $k$th point by

$$n_k(t+1) = n_{dk}(t+1)\lambda_{dk}(t+1) + n_{sk}(t+1)\lambda_{sk}(t+1) \quad (16)$$

where $\lambda_{dk}(t+1)$ and $\lambda_{sk}(t+1)$ denote the combination ratios for the diffuse and specular components.

Since the structure of the proposed neural networks is like a mirror in the center layer, the update rule for the weights between Layers 2 and 3 of the two subnetworks denoted as $W_d$ and $W_s$ can be calculated by the least square method. Hence, $W_d$ and $W_s$ at time $t+1$ can be calculated by

$$W_d(t+1) = (V_d(t+1)^T V_d(t+1))^{-1} V_d(t+1)^T$$

$$W_s(t+1) = (V_s(t+1)^T V_s(t+1))^{-1} V_s(t+1)^T \quad (17)$$

where $V_d(t+1)$ and $V_s(t+1)$ denote the weights betweens the output and central layers of the two subnetworks for the diffuse and specular components, respectively.

Additionally, for fast convergence, the learning rate η of the neural network is adaptive in the updating process. If the current error is smaller than the errors of the previous two iterations, then the current direction of adjustment is correct.

Thus, the current direction should be maintained, and the step size should be increased, to speed up convergence. By contrast, if the current error is larger than the errors of the previous two iterations, then the step size must be decreased because the current adjustment is wrong. Otherwise, the learning rate does not change. Thus, the cost function $E_T$ could reach the minimum quickly and avoid oscillation around the local minimum. The adjustment rule of the learning rate is given as follows:

If (Err (t-1) > Err (t) and Err (t-2) > Err (t) )
η(t+1)= η(t) + ξ ,
Else If (Err (t-1) < Err (t) and Err (t-2) < Err (t) )
η(t+1)= η(t) − ξ , where ξ is a given scalar.
Else η(t+1)= η(t)

## IV. EXPERIMENT RESULTS AND DISCUSSION

Yale Database has been used in this project and from this database two datasets have been considered for training [5]. In the first dataset a person with fixed pose, 5 images were selected with 5 different illuminant direction. In the second dataset five people with fixed pose, 3 images were taken with 3 different illuminant direction. Illuminant directions chosen for datasets are different.

After the text edit has been completed, the paper is ready for the template. Duplicate the template file by using the Save As command, and use the naming convention prescribed by your conference for the name of your paper. In this newly created file, highlight all of the contents and import your prepared text file. You are now ready to style your paper; use the scroll down window on the left of the MS Word Formatting toolbar.

### A. Preprocessing

In the preprocessing stage, the images were cropped into 64X64 pixels. It was made into a single vector of size 4096. Therefore the first database is of size 3X4096 and the second database is of size 9X4096.

## B. Training Results

The neural network was implemented and the training of the network was started with the first dataset. The results of the training done so far are given below.

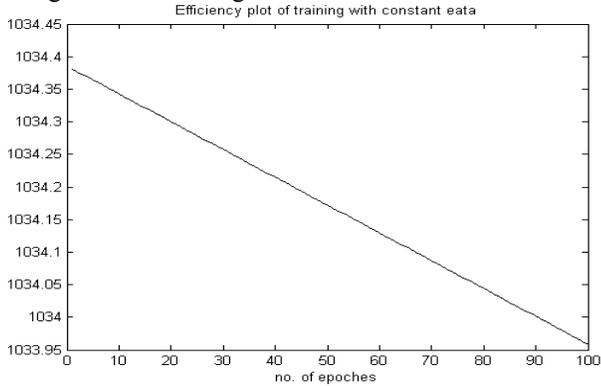

Figure 3 Efficiency plot with constant eata

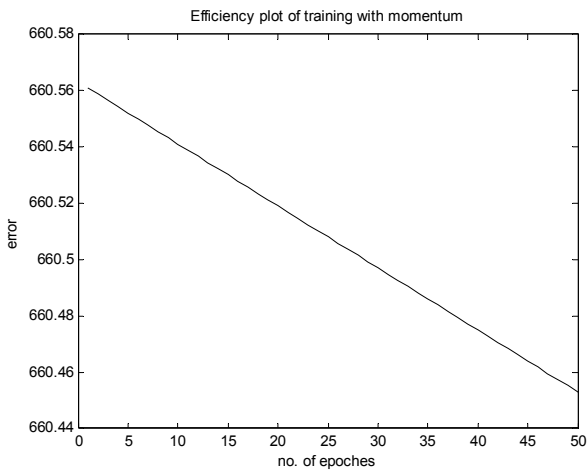

Figure 4 Efficiency plot with momentum

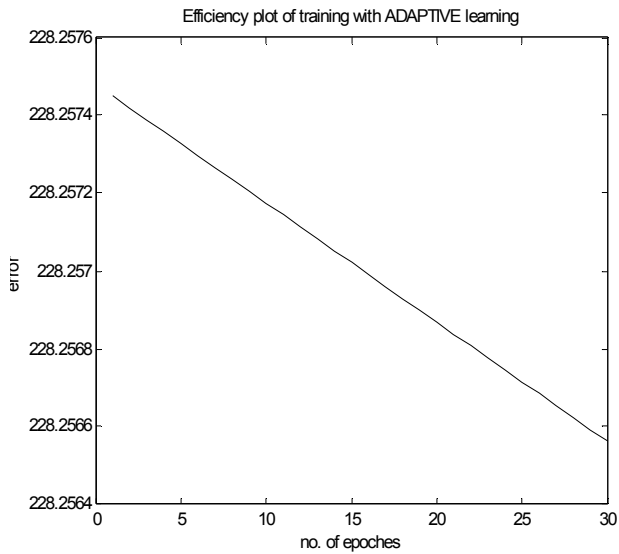

Figure 5 Efficiency plot with adaptive eata

First the training was started with a fixed learning constant. It was working fine for higher values of error and later the convergence of error was very slow. Then the training was done applying momentum. It reduced the error to some lower value and it started degrading very slowly. Then adaptive learning method was employed which showed that the convergence can happen faster. Thus adaptive learning method is a faster method in error back propagation algorithm.

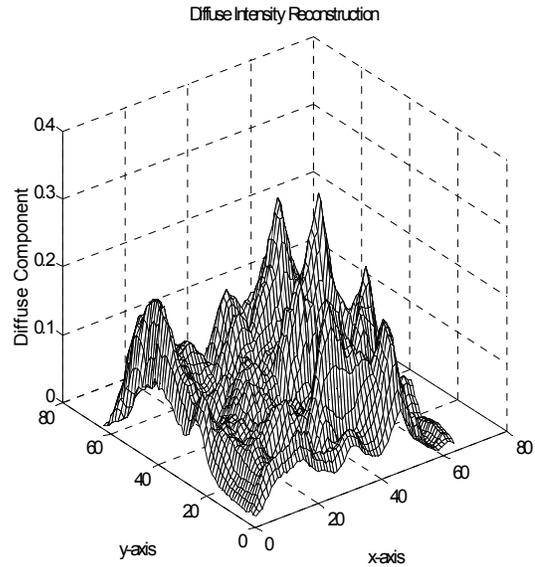

Figure 6 Reconstruction with Diffuse Component

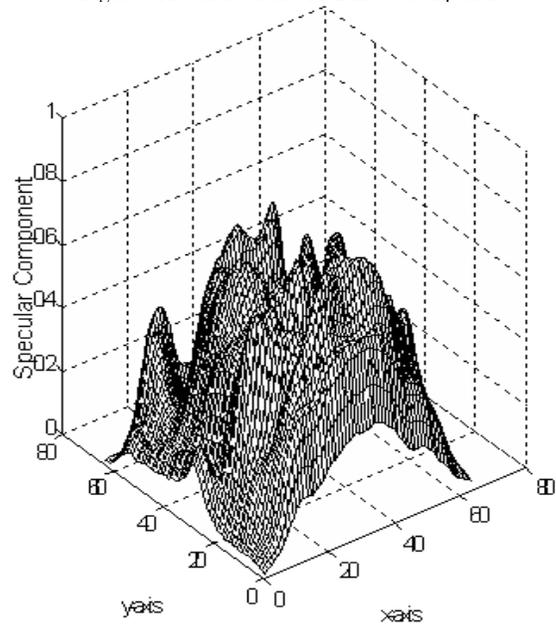

Figure 7.Reconstruction with Specular Component

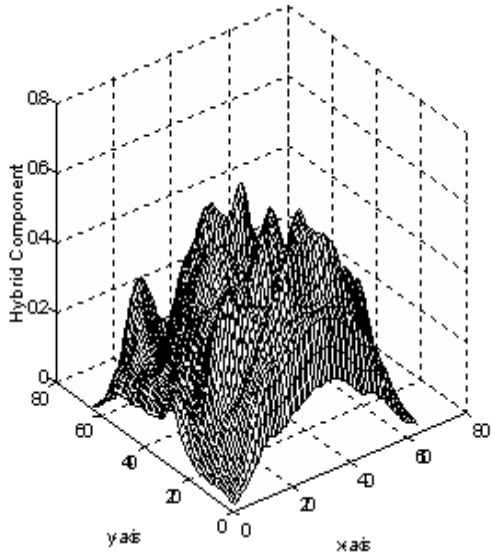

Figure 8.    Reconstruction with Hybrid Component

## V.    CONCLUSION

In this paper a novel 3-D image reconstruction approach which considers both diffuse and specular components of the reflectance model simultaneously has been proposed. Two neural networks with symmetric structures were used to estimate these two components separately and to combine them with an adaptive ratio for each point on the object surface. This paper also attempted to reduce distortion caused by variable albedo variation by dividing each pixel's intensity by corresponding albedo value. Then, these intensity values were fed into network to learn the normal vectors of surface by back-propagation learning algorithm. The parameters such as light source and viewing direction can be obtained from the neural network. The normal surface vectors thus obtained can then be applied to 3-D surface reconstruction by  integration method.

Shalini Bhatia was born on August 08, 1971. She received the B.E. degree in Computer Engineering from Sri Sant Gajanan Maharaj College of Engineering, Amravati University, Shegaon, Maharashtra, India in 1993, M.E. degree in Computer Engineering from Thadomal Shahani Engineering College, Mumbai, Maharashtra, India in 2003. She has been associated with Thadomal Shahani Engineering College since 1995, where she has worked as Lecturer in Computer Engineering Department from Jan 1995 to Dec 2004 and as Assistant Professor from Dec 2004 to Dec 2005. She has published a number of technical papers in National and International Conferences. She is a member of CSI and SIGAI which is a part of CSI.

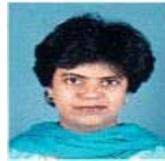

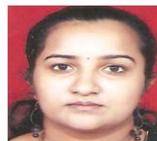

Vincy Elizabeth Joseph was born on February 5, 1982. She received the B.E. degree in Electronics and Communication Engineering from College of Engineering, Kidangoor, Cochin University of Science and Technology, Cochin, Kerala. She is pursuing M.E. degree in Computer Engineering from Thadomal Shahani Engineering College, Mumbai, Maharashtra, India.She is working with St.Francis Institute of Technology, Borivli (W) Mumbai from the year 2004 to 2005 as Lecturer in Electronics and Telecommunication Department and from the year 2005 as Lecturer in Computer Engineering Department. Her research interests include Image Processssing, Neural Networks, Data Encryption and Data Compression